\documentclass[conference]{ieeeconf}  
\IEEEoverridecommandlockouts       

\usepackage[utf8]{inputenc}
\usepackage[T1]{fontenc}
\usepackage{graphicx}
\usepackage{color}
\usepackage[top=1in, bottom=1.2in, left=0.8in, right=0.8in]{geometry}
\usepackage{comment}
\usepackage{caption}
\usepackage{subcaption}
\usepackage{graphicx}
\usepackage{import}
\usepackage{amsmath}
\usepackage{amssymb}
\usepackage{url}
\usepackage[colorinlistoftodos]{todonotes}
\newcommand{\TODO}[1]{{\color{red}#1}}

\newcommand{\diego}[1]{{\color{olive}#1}}

\renewcommand{\baselinestretch}{0.975}

\newcommand{\fixme}[1]{\todo[size=\footnotesize, color=red!40]{#1}}

\begin{document}

\title{\bf A Novel Low-Cost, Recyclable, Easy-to-Build Robot Blimp For Transporting Supplies in Hard-to-Reach Locations}

\author{
Karen Li${}^{\ddag}$, Shuhang Hou${}^{\ddag}$, Matyas Negash${}^{\ddag}$, Jiawei Xu${}^{\ddag}$, 
\\
Edward Jeffs${}^{\ddag}$, Diego S. D'Antonio${}^{\ddag}$, 
and David Salda\~{n}a${}^{*}$
\\
Lehigh University, Bethlehem, PA USA
\\
Email – saldana@lehigh.edu
\thanks{${}^{\ddag}$K. Li, S. Hou, M. Negash, J. Xu, Edward Jeffs, D. D'Antonio, and ${}^{*}$D. Salda\~{n}a are with the Autonomous and Intelligent Robotics Laboratory --AirLab-- at Lehigh University, PA, USA: $\{$\texttt{kkl224, shh420, mkn325, jix519, elj221, diego.s.dantonio, saldana$\}$@lehigh.edu}}
\thanks{The authors appreciate the help and support during the experiments of Bingxu Zhao, Hanqing Qi, Tinotenda Chibvuri, and Leonardo Santens at Lehigh University.}
\thanks{We would also like to thank Dr. Khanjan Mehta for advice and discussions about innovative technology and social impact.}
\thanks{The authors gratefully acknowledge the support from the Office of Naval Research, Grant  N00014-23-1-2535, and the Office of Creative Inquiry at Lehigh University.}
}

\maketitle
\thispagestyle{empty}
\pagestyle{empty}


\begin{abstract}
Rural communities in remote areas often encounter significant challenges when it comes to accessing emergency healthcare services and essential supplies due to a lack of adequate transportation infrastructure. The situation is further exacerbated by poorly maintained, damaged, or flooded roads, making it arduous for rural residents to obtain the necessary aid in critical situations. Limited budgets and technological constraints pose additional obstacles, hindering the prompt response of local rescue teams during emergencies. The transportation of crucial resources, such as medical supplies and food, plays a vital role in saving lives in these situations. In light of these obstacles, our objective is to improve accessibility and alleviate the suffering of vulnerable populations by automating transportation tasks using low-cost robotic systems. We propose a low-cost, easy-to-build blimp robot (UAVs), that can significantly enhance the efficiency and effectiveness of local emergency responses. 
\end{abstract}
\smallskip


\section{Introduction}

In rural communities located in remote areas, the absence of transportation infrastructure presents considerable obstacles in the population's access to emergency healthcare, first aid kits, and vital supplies during times of crisis \cite{Syed_Gerber_Sharp_2013}. This issue is particularly pronounced in countries like the Philippines, which has a unique and challenging geographical layout \cite{1381948}. The scattered nature of the Philippine islands and the rugged terrain make it exceptionally difficult to establish well-developed road networks, resulting in limited accessibility to emergency services and supplies. Adding to the complexity, the country faces a high frequency and intensity of natural disasters due to its geographical layout, with some estimations placing 60\% of its land area and 74\% of its population as exposed to numerous hazards \cite{Lloyd_Gray_Healey_Opdyke_2022}. In the aftermath of a disaster, debris, fallen trees, or floodwaters can block roads and hinder transportation. This restriction on road access significantly affects emergency response efforts, particularly the timely delivery of essential supplies. Consequently, addressing these transportation challenges in rural areas becomes crucial. It requires concerted efforts to implement strategies that enhance accessibility to emergency services and supplies. 
\smallskip


\begin{figure}
    \centering
    \includegraphics[width=\linewidth]{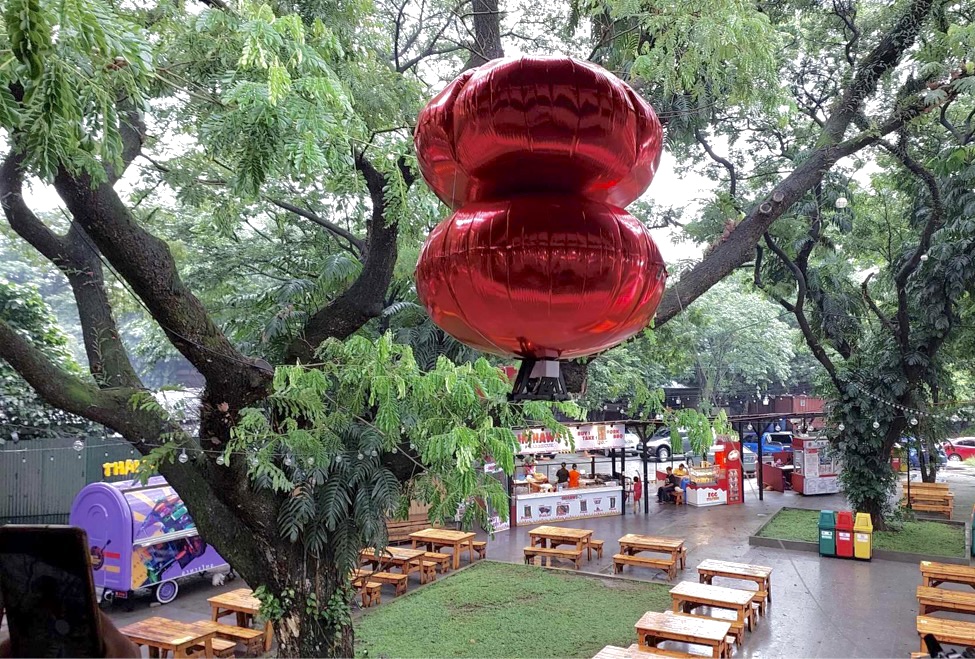}
    \caption{Current prototype capable of lifting objects weighing 50\;g and flying, tested in an outdoor environment with a wind speed of up to 2 m/s. The blimp was flying on the campus of the University of the Philippines.}
    \label{fig:prototype}
    \vspace{-0.8em}
\end{figure}

With the advent of modern technology, there is an increasing recognition of the crucial role science and technology play in humanitarian development \cite{development2022}. To address the issue of inaccessibility in rural communities, innovative tools and methodologies are needed to effectively tackle these problems. Unmanned Aerial Vehicles (UAVs) have emerged as valuable assets for improving accessibility due to their unique capabilities and versatility \cite{RHI2023}. Leveraging the advantages of drone technology can help to overcome the limitations imposed by rugged terrains and the absence of well-developed transportation infrastructure. 

The drone delivery space has experienced remarkable growth and innovation in recent years \cite{McKinsey_2022}, spearheaded by industry leaders such as Zipline, Matternet, and Flytrex. These companies are at the forefront of revolutionizing the transportation of goods, particularly in remote or inaccessible areas. Among them, Zipline stands out as the most promising player, having successfully executed over 450,000 deliveries \cite{Korosec_2022}. Zipline has established a proven market model by specializing in the delivery of crucial medical supplies, including blood, ointments, and medicine, in Rwanda. Their choice of operating fixed-wing drones allows them to achieve higher speeds, reaching up to 70 mph, and enables efficient long-range flights \cite{mpetrova92_2018}. To launch their drones, Zipline utilizes a catapult launcher, ensuring swift takeoff. Their aircraft releases its payload through a hatch, relying on a parachute for the payload to descend to the ground. While this approach works well in areas with ample landing zones and extensive infrastructure such as launchers, it poses challenges when the delivery takes place on the side of a mountain or a small island. Quadcopters offer an alternative solution, which rely solely on motors for lift generation, as opposed to utilizing airflow through the wings. Consequently, quadcopters have significantly shorter flight times, typically less then 50\% of what fixed-wing counterparts can achieve. While UAVs have great potential, their high initial cost and ongoing maintenance expenses may limit their widespread adoption, especially in rural communities facing budget constraints. In order to address this issue, cost-effective tools and methodologies are required to better handle these situations.




In contrast to fixed-wing aircraft and quadcopters, blimps utilize lighter-than-air gases within their body to generate the majority of their lifting force. This unique characteristic offers several advantages over their counterparts. By incorporating blimp technology into drone delivery systems, we can overcome the limitations faced by fixed-wing aircraft and quadcopters. However, there is very limited research and development in the field of drone delivery utilizing blimp technology to address the challenges including fuel efficiency, noise levels, and flight time. To bridge this gap, we propose a solution that leverages blimp robots for drone applications in realistic outdoor environments. By combining the advantages of blimps with robot capabilities, we aim to create a feasible and sustainable approach to a scalable drone delivery with the following advantages:

\paragraph{Extended Flight Time} Blimp robots offer the advantage of longer flight durations compared to traditional drones that rely on mechanical components like wings or rotors. Instead of solely relying on batteries with limited capacity, blimps utilize the buoyancy provided by lighter-than-air (LTA) gas to stay afloat for extended periods, which help us accomplish tasks that require prolonged operation without interruption. 
\paragraph{Maximized Payload} The design of blimps allows for the carriage of larger and heavier payloads, making them ideal for transporting essential resources such as medical supplies to disaster-affected areas, which contributes to their significance in facilitating deliveries in various applications.
\paragraph{Ease of Takeoff and Land} Blimps have a distinct advantage when it comes to takeoff due to their lighter-than-air nature. Unlike fixed-wing aircraft, blimps require minimal runway or takeoff infrastructure. They have the ability to take off and land vertically, smoothly and steadily \cite{Petrescu2017}. The ease of takeoff adds to the versatility and adaptability of blimps, making them a valuable asset in diverse operational environments.
\paragraph{Collision Resilience} The soft and flexible nature of the balloon envelope in blimps allows them to absorb impacts with obstacles or structures, making them well-suited for operating in cluttered or confined environments where the risk of collisions and potential damage is higher \cite{doi:10.1177/1756829317705326,huang2019duckiefloat}. Their collision resilience enhances safety and reliability, enabling them to continue operating and providing valuable support in scenarios such as disaster response missions.

We present in this paper a novel solution that utilizes drone technology to enhance accessibility in areas grappling with transportation barriers, such as the Philippines. We discuss how blimp robots can overcome the limitations of current technology and provide a more feasible and practical solution for emergency response in resource-constrained areas. By using affordable and readily available materials, we introduce a blimp robot model that can be constructed without significant financial burden, making it accessible to communities with limited resources. We also discuss the important challenges that need to be addressed to enable the seamless use and adoption of blimp robots, which include technical considerations and operational limitations.

\smallskip

\section{Our Approach}
The blimp was meticulously designed to prioritize simplicity and feasibility, and it has three primary components:
\subsubsection{Flight Controller} The propulsion system of the blimp utilizes the low-cost microcontroller ESP32-S3 as flight control board. Despite its compact size, weighing just $1.7 \ g$ and having a small dimension of $21 \times 17.5 \ mm$. The processing power of ESP32-S3 is sufficient to control 2 servos and 2 motors. Each motor is mounted on a servo to assemble a bi-copter mechanism.
The simplicity of our design, using minimal actuation support, reduces complexity and the risk of potential points of failure, provides adequate support for takeoff and landing tasks.

\subsubsection{Foldable chassis} 
\begin{figure}
    \centering
    \includegraphics[height=0.6\linewidth]{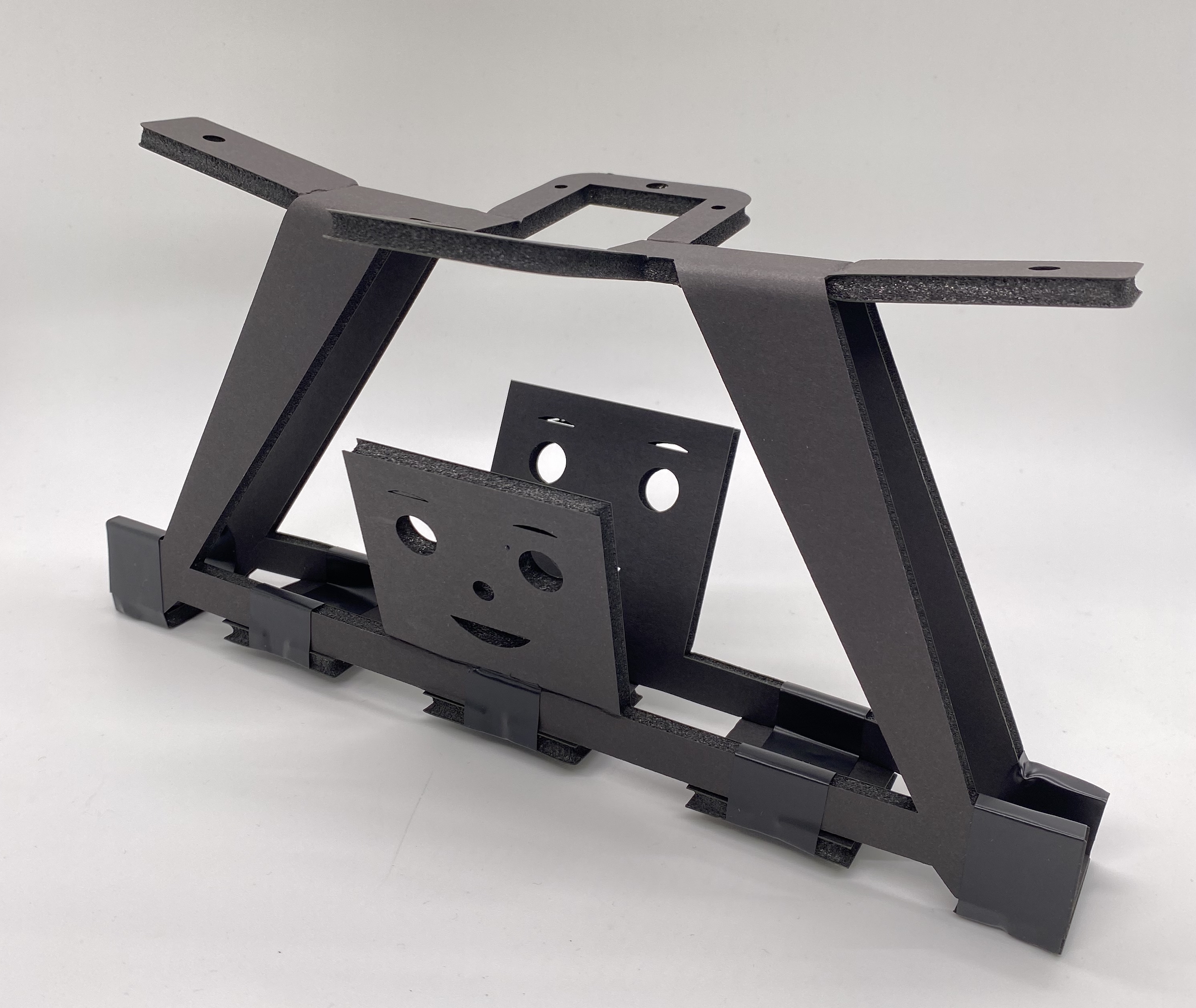}
    \caption{Foldable chassis used to hold the components}
    \label{fig:chassis}
    \vspace{-0.4em}
\end{figure}

We introduce a foldable chassis design utilizing paper materials for the robot blimp's structure, offering an efficient and cost-effective construction solution. Drawing inspiration from principles found in origami and bridges, we have engineered the paper material to exhibit sufficient strength for securely holding all components in place, while also providing the capability to bear additional weight. Crutially, the internal structure can support forces up to $5 \ kg$. To implement our approach, we employ a laser cutter or a scalpel to precisely cut a foam core board, which serves as the basis for the chassis. Upon cutting, the chassis can be effortlessly folded, forming the desired framework for the blimp. (See Fig. \ref{fig:chassis}). 
\subsubsection{Helium Balloon} The blimp's main body consists of a helium-filled balloon made from Mylar material. The balloon takes on the shape of an ellipsoid, providing stability and aerodynamic performance. The total volume of the ellipsoid-shaped balloon is $0.125 \ m^3$, and it is filled with industry-grade helium, guaranteeing a minimum helium concentration of $99\%$. The blimp generates up to $65 \ g$ buoyancy 
, providing sufficient lifting force to transport all necessary components as well as any additional payloads, if required.


The blimp's design features practicality and reliability, incorporating several noteworthy aspects:

\paragraph{Balloon Attachment} To ensure a secure connection between the balloon and the foldable chassis, we employ two sewing elastic bands forming an `X' configuration (See Fig.~\ref{fig:prototype}). These elastic bands offer flexibility and can accommodate the expansion and contraction of the balloon as needed. The tension provided by the elastic bands keeps the chassis securely in place, eliminating the need for tape or other adhesives. Moreover, this balloon attachment design facilitates the effortless replacement of balloons. The use of sewing elastic bands exemplifies a practical and adaptable method for firmly attaching the balloon to the structure, allowing for a reliable and adjustable connection.

\begin{figure}[t]
    \centering
    \includegraphics[height=0.6\linewidth]{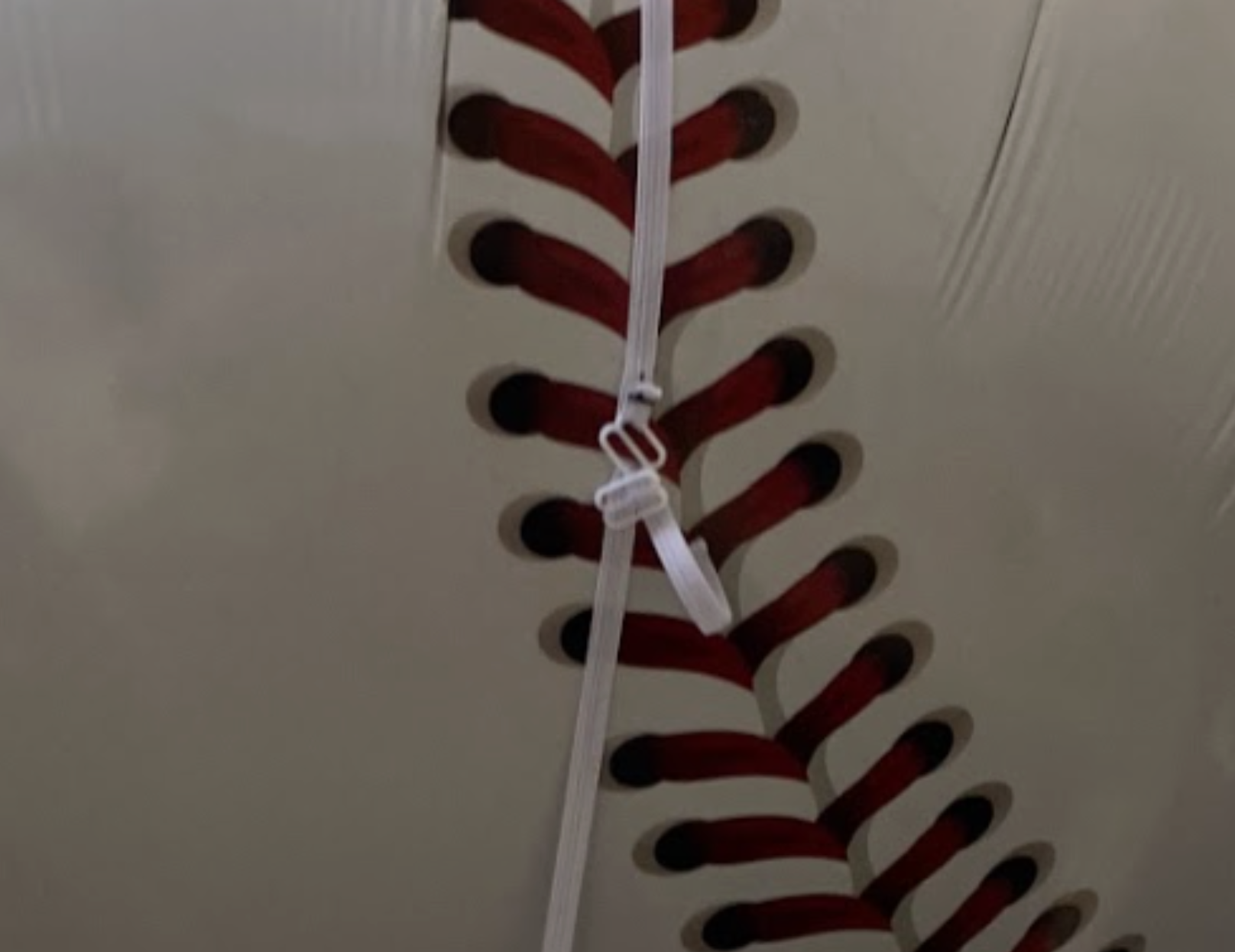}
    \caption{Plastic clips utilized to adjust tightness by sliding along the bands.}
    \label{fig:bands}
    \vspace{-0.4em}
\end{figure}

\paragraph{Adjustable Fastening} To further enhance the flexibility and adaptability of the system, plastic clips are utilized to tighten the sewing elastic bands (See Fig.~\ref{fig:bands}). These clips enable fine-tuning of the tightness, providing precise control over the attachment of the balloon to the chassis. This adjustability is particularly beneficial in situations where the helium balloon may lose air or deflect, as it allows for prompt tightening to maintain the structural integrity of the blimp robot. The use of plastic clips offers a convenient and effective means of adjusting the tension in the elastic bands. By easily sliding the clips along the bands and securing them in place, the desired level of tightness can be achieved, thus ensuring that the balloon remains securely fastened to the chassis. The incorporation of plastic clips in the attachment mechanism adds to the overall flexibility and adaptability of the system. 


\begin{table}[b]
    \centering
    \begin{tabular}{|c|c|c|}
    \hline
        Components & QTY & C/U (USD)\\    
        \hline \hline
        XIAO ESP32 S3 & 1 & 5\\
        Sensor Board & 1 & 4.88\\
        Brushed motors & 2 & 2.4\\
        Servo motor & 2 & 0.8\\
        Motor driver & 1 &1.49\\
        Paper frame & 1 & 2\\   
        Balloon & 2 & 2\\
        Helium & 124~liter & 3.65\\
    \hline
    Total & 1 & 27.42\\
    \hline
    \end{tabular}
    \caption{Total cost of a blimp robot.}
    \label{tab:my_label}
\end{table}


\paragraph{Foldable Chassis Positioning} Based on the blimp components design, the vehicle naturally tries to stay horizontal. The rigid placement of the object underneath the top of chassis body beneath the intersection of the balloon's major axes is a key design feature that emphasizes the blimp's low center of mass and enhances its structural integrity. The low center of mass plays a crucial role in ensuring the blimp's natural stability during flight, as it helps to counterbalance any external forces or disturbances that may affect the vehicle. By positioning the motors, servos, and associated components in a rigid configuration inside the chassis, such a design minimizes potential vibrations and stresses, promoting safe and controlled operation.


\begin{figure}
    \centering
    \includegraphics[height=0.66\linewidth]{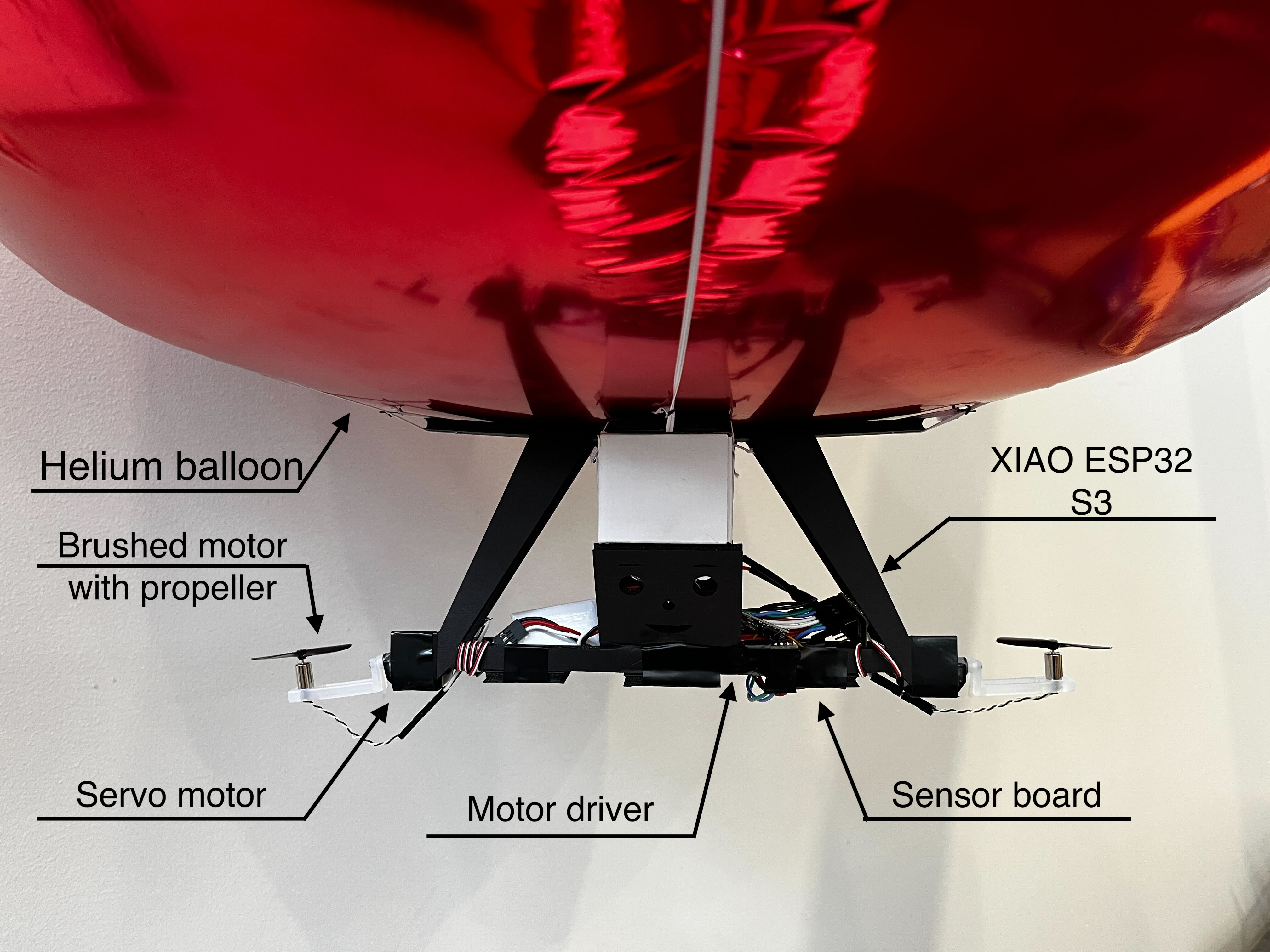}
    \caption{Front view of the vehicle.}
    \label{fig:front}
    \vspace{-0.8em}
\end{figure}


\smallskip
\section{Use Cases}
Drone delivery has numerous application in places lack of connectivity as it addresses various challenges and improves efficiency in many different sectors. Thus, the opportunity route for utilization of drones is vast and diverse. Here are some possible uses cases:

\paragraph{Rural Connectivity} In less developed countries like the Philippines, numerous remote and underserved rural areas suffer from limited access to essential facilities. This lack of physical connectivity creates barriers for rural residents, hindering their ability to reach crucial services such as healthcare facilities, schools, and markets. Without well-maintained roads or reliable public transportation systems, individuals face significant challenges in accessing specialized medical care or attending educational institutions in nearby towns or cities. However, the introduction of drone technology can address these limitations by providing efficient and cost-effective transportation solutions, ultimately generating educational impact and promoting knowledge transfer. By implementing drone technology, not only can essential goods, medical supplies, and agricultural products be transported to remote rural areas in a swift and efficient manner, but the technology itself can also have transformative educational benefits. Drones equipped with user-friendly interfaces and simplified operation mechanisms can empower non-expert individuals to become familiar with cutting-edge technology. This exposure to drones not only facilitates the transport of goods but also serves as a catalyst for knowledge transfer and skill development. As more people become acquainted with drone maneuvering, the technology can be leveraged to address a wider range of challenges faced by rural communities. By gradually expanding the scope of drone applications, rural residents gain exposure to new opportunities and possibilities, fostering a culture of innovation and problem-solving.

\paragraph{Emergency Response} One useful application is the delivery of medical supplies during emergency response. In countries like the Philippines, where road access can be limited, delivering medical supplies for emergency response can be challenging and time-consuming. The geographical features of these countries make it difficult to establish and maintain efficient transportation networks, especially in areas with limited road infrastructure. However, drones can quickly transport essential medical supplies, such as medications, vaccines, test strips, or diagnostic samples, to inaccessible areas. They can bypass traffic congestion and rough terrain, ensuring the timely delivery of critical resources to emergency response teams. This can be particularly valuable during emergency situations where immediate medical attention is required. The establishment of a drone-based transportation network can enhance the coordination and connectivity between medical facilities. The integration of blimps into the existing infrastructure enables efficient distribution of resources during emergencies.


\begin{figure}
    \centering
    \includegraphics[height=0.66\linewidth]{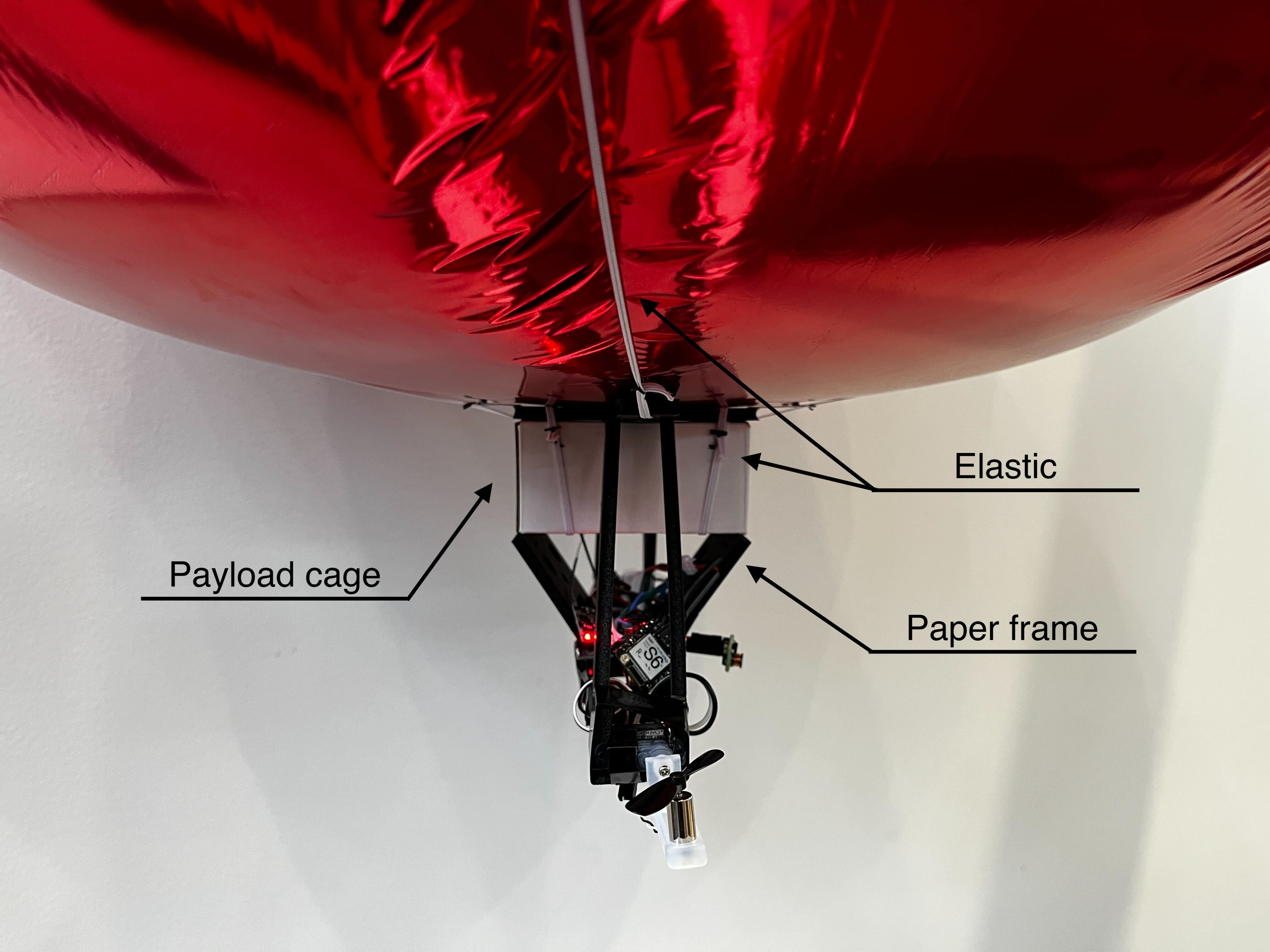}
    \caption{Side view of the vehicle.}
    \label{fig:side}
    \vspace{-0.8em}
\end{figure}


\paragraph{Disaster Relief} In the aftermath of a disaster, the disruption caused by debris, fallen trees, or floodwaters often obstructs roads and severely hampers transportation. This poses significant challenges for both affected individuals in need of assistance and response teams attempting to coordinate their efforts. To mitigate these difficulties and minimize losses during the crucial post-disaster response phase, the implementation of efficient management practices integrated with digital technologies becomes imperative. In this regard, the utilization of drone delivery approaches emerges as a game-changer in disaster relief efforts. Drones, with their remarkable capabilities, can play a pivotal role in swiftly and effectively transporting essential emergency supplies to impacted communities. These supplies encompass vital resources such as food, water, blankets, and hygiene kits. By leveraging drone delivery, these provisions can be rapidly transported, bypassing disrupted transportation infrastructure and reaching otherwise inaccessible areas in a timely manner.


\smallskip
\section{Model}
\subsection{Dynamics}
\begin{figure}[t]
    \centering
    \includegraphics[width=\linewidth]{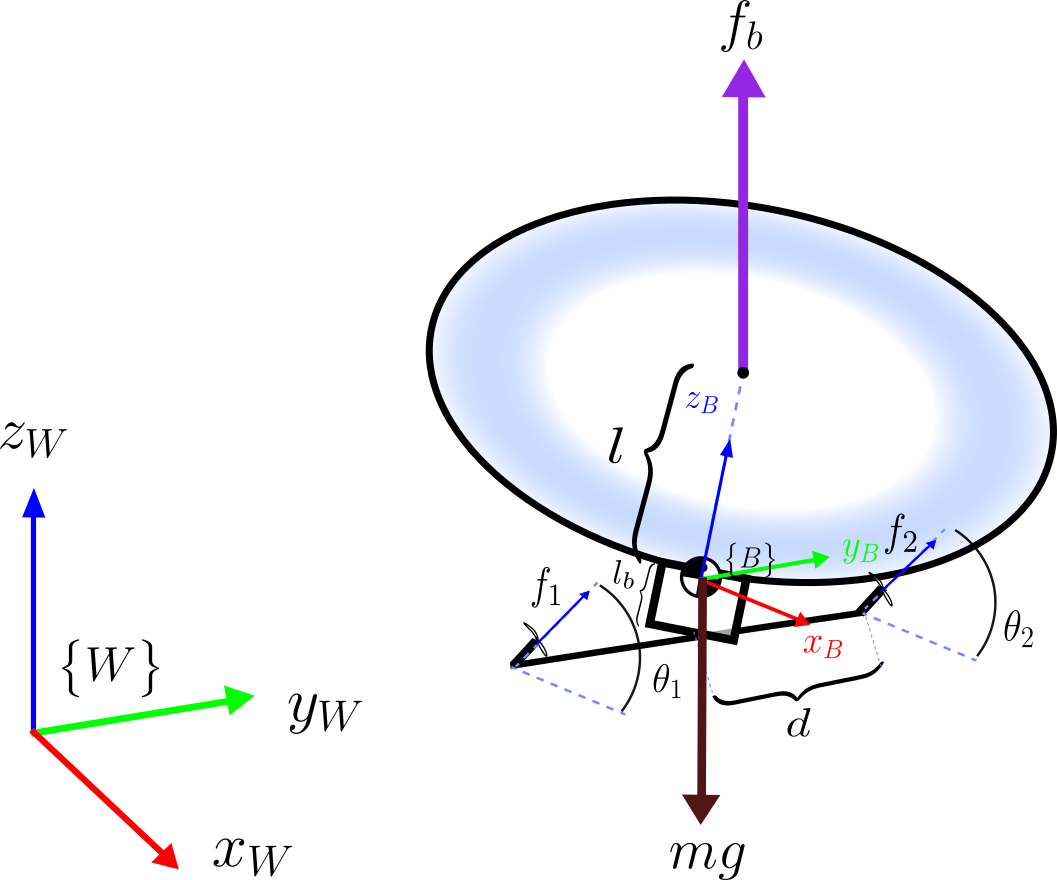}
    \caption{The abstracted model of our blimp, where the dowel points represents the center of mass of the blimp.}
    \vspace{-0.4em}
    \label{fig:blimpmodel}
\end{figure}

We define the world reference frame as a fixed frame, denoted by $\{W\}$. The blimp has a body frame $\{B\}$ whose origin is at the center of mass (COM). Its $x$-axis points toward the front of the blimp, and its $z$ axis points upward, as shown in Fig.~\ref{fig:blimpmodel}. A pair of vectorized thrusters actuate the blimp to achieve rotational and translational motion. The vectorized thrusters are mounted at both end of a support arm placed beneath the balloon in the fashion of a bicopter~\cite{7748042}, both keeping a distance of $d$ from $z_B$. The support arm is in parallel with the $y$-axis of $\{B\}$ with a distance of $l_b$ below the blimp COM. Therefore, we denote their mounting positions in $\{B\}$ as $\boldsymbol{p}_1 = \left[0, -d, l_b\right]^\top$ and $\boldsymbol{p}_2 = \left[0, d, l_b\right]^\top$, respectively. Each vectorized thruster consists of a micro servo and a rotor. The thrust forces of the rotors are $f_1$ and $f_2$, respectively, and the rotation angles of the servos in the direction of $y_B$ are $\theta_1$ and $\theta_2$, respectively. At rest, i.e., $\theta_i = 0$, the force vector of the $i$-th thruster aligns with $x_B$.

The translation and rotation from $\{W\}$ to $\{B\}$, $\boldsymbol{r}\in\mathbb{R}^3$ and $\boldsymbol{R}$, respectively, describe the position and orientation of the blimp. The rotation matrix, $\boldsymbol{R}\in\mathsf{SO(3)}$, is in the special orthogonal group of dimension $3$, which means $\det{{\boldsymbol{R} = 1}}$ and $\boldsymbol{R}^{-1} = \boldsymbol{R}^\top$. $\boldsymbol{R}$ can be converted from Euler angles with
\begin{equation}
    \boldsymbol{R} = \begin{bmatrix}
        c\psi c\theta-s\phi s\psi s\theta & -c\phi s\psi & c\psi s\theta+c\theta s\phi s\psi\cr c\theta s\psi+c\psi s\phi s\theta & c\phi c\psi & s\psi s\theta-c\psi c\theta s\phi\cr -c\phi s\theta & s\phi & c\phi c\theta
    \end{bmatrix}
    \label{eq:rotm}
\end{equation}
where $\phi, \theta, \text{ and } \psi$ represent the corresponding roll-pitch-yaw Euler angles in the direction of $x_W, y_W$, and $z_W$, respectively, and $c\theta$ and $s\theta$ denote $\cos\theta$ and $\sin\theta$, respectively, similarly for $\phi$ and $\psi$.

We use Newton-Euler equation to describe the dynamics of the blimp,
\begin{align}
    m\boldsymbol{\Ddot{r}} = &\boldsymbol{Rf} + \boldsymbol{f}_e,\label{eq:newton}\\
    \boldsymbol{J\dot{\omega}} + \boldsymbol{\omega}\times\boldsymbol{J\omega} = & \boldsymbol{\tau} + \boldsymbol{\tau}_e,\label{eq:euler}
\end{align}
where the net force and torque vectors generated by the thrusters in $\{B\}$ are 
\begin{align}
    \boldsymbol{f} &= \sum_{i=1}^{2}f_i\left[\cos{\theta_i}, 0, \sin{\theta_i}\right]^\top,\label{eq:forceB}\\
    \boldsymbol{\tau} &= \sum_{i=1}^{2}f_i\left(\boldsymbol{p}_i\times\left[\cos{\theta_i}, 0, \sin{\theta_i}\right]^\top\right).\label{eq:torqueB}
\end{align}

The external force from gravity and the balloon buoyancy is $\boldsymbol{f}_e = \left[0, 0, f_b - mg\right]^\top$, where $g$ is the gravitational acceleration and $m$ is the total mass of the blimp. The external torque from the buoyancy is $\boldsymbol{\tau}_e = \left(\boldsymbol{R}\left[0, 0, l\right]\right)\times\left[0, 0, f_b\right]^\top$ and $\boldsymbol{J}$ is the matrix of inertia moment.~\eqref{eq:newton} shows that the combined forces of gravity, buoyancy, and rotor thrust bring a linear acceleration to the blimp, causing it to translate.~\eqref{eq:euler} shows that the combined torque of gravity, buoyancy, and rotor thrust bring an angular acceleration to the blimp, causing it to rotate. 


\subsection{Control}
\subsubsection{Manual Control}
    We use a joystick to provide manual control input for the blimp. Under manual control, the motion is egocentric, which means that users only control the force and torque in $\{B\}$, as shown in~\eqref{eq:forceB} and~\eqref{eq:torqueB}. Similarly to controlling a differential drive ground vehicle~\cite{lynch2017modern} using a joystick controller, the user determines the linear motion of the blimp in the $xz$-plane and the angular motion in yaw, marked by forces $f_x$, $f_z$ which are the first and third elements in~\eqref{eq:forceB} and torque $\tau_z$ which is the third element in~\eqref{eq:torqueB}, respectively. Since the blimp has four actuators, namely, two rotors and two servos, we further allow the user to provide a desired $\tau_x$ which is the first element in~\eqref{eq:torqueB} for stable roll operation.
    Solving for the actuator inputs $f_1, f_2, \theta_1$, and $\theta_2$ with given $f_x, f_z, \tau_x$, and $\tau_z$, we obtain 
    \begin{align}
        f_1 &= \sqrt{f_{1x}^2 + f_{1z}^2},\nonumber\\
        f_2 &= \sqrt{f_{2x}^2 + f_{2z}^2},\nonumber\\
        t_1 &= \arctan\frac{f_{1z}}{f_{1x}},\nonumber\\
        t_2 &= \arctan\frac{f_{2z}}{f_{2x}},\nonumber
    \end{align}
    where $f_{1x} = \frac{1}{2}(f_x - \frac{\tau_z}{d}),
        f_{2x} = \frac{1}{2}(f_x + \frac{\tau_z}{d}),
        f_{1z} = \frac{1}{2}(f_z + \frac{\tau_x}{d}),\text{ and }
        f_{2z} = \frac{1}{2}(f_z - \frac{\tau_x}{d})$.
    
\subsubsection{Autonomous control}
    We control the blimp to go to a desired position $\boldsymbol{r}^d$ by applying a Proportional-Integral-Derivative (PID) control~\cite{unbehauen2009control} to obtain the desired force in $\{W\}$
    \begin{equation}
        \boldsymbol{f}^d = \boldsymbol{K}_p\left(\boldsymbol{r}^d - \boldsymbol{r}\right) + \boldsymbol{K}_d\left(\boldsymbol{\dot r}^d - \boldsymbol{\dot r}\right) + \boldsymbol{K}_i\int\left(\boldsymbol{r}^d - \boldsymbol{r}\right)dt,
    \end{equation}
    where $\boldsymbol{K}_p$, $\boldsymbol{K}_d$, and $\boldsymbol{K}_i$ are the proportional, derivative, and integral gain matrices. To generate the force, the blimp needs to consider its orientation, and transform the desired force into $\{B\}$, $\boldsymbol{f}_b = \boldsymbol{R}^\top\boldsymbol{f}^d$. Similarly to manual control, the blimp achieves the desired motion in its $xz$-plane by taking the first and third elements of $\boldsymbol{f}^d$ as $f_x$ and $f_z$, then compensates the desired motion in $y_B$ directions by converting the second element of $\boldsymbol{f}_b$, $f_y$ into the desired torque in yaw, $\tau_z = K_\psi\arccos{\frac{f_y}{\Vert\boldsymbol{f}^d\Vert}}$, where $K_\psi$ is a gain coefficient for yaw.

\section{Experiments}
\begin{figure}
    \centering
    \includegraphics[width=0.9\linewidth]{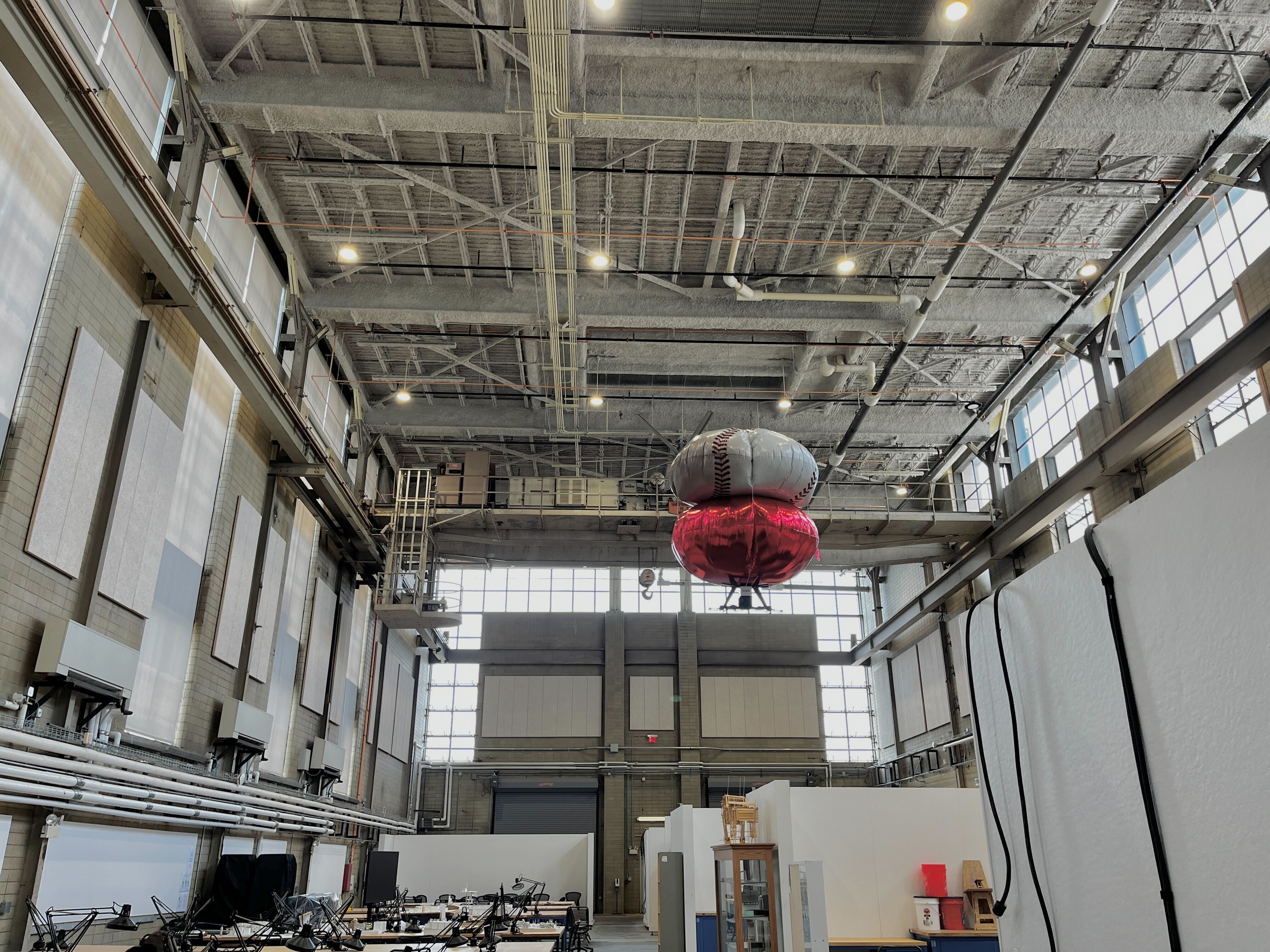}
    \caption{A blimp robot carrying an object weighing $50 \ g$ flying in a high-bay area with air-conditioners on the side generating random turbulence, with velocities reaching up to $0.4 \ m/s$, simulating a realistic environment.}
    \label{fig:highbay}
    \vspace{-0.3em}
\end{figure}

To evaluate the performance and limitations of our vehicle design and control system, we conducted a series of experiments using the prototype blimp robot, as described in Section V, with manual control via a joystick. These experiments aim to evaluate the blimp's ability to transport a small box weighing $50$ grams from point A to point B, thereby validating the effectiveness of the blimp robot design and controller. The experiments involve subjecting the blimp robot to various conditions, including turbulent environments and potential collisions. 

\subsection{Flying in environments with low turbulence}
This evaluation involves transporting objects from one point to another, covering a distance of 60 meters in a realistic environment where turbulence is present (See Fig. 6). To simulate turbulent conditions, we introduce random turbulence generated by air conditioners placed in the high-bay area. The velocity of the turbulence reaches up to 0.4 meters per second, providing a realistic scenario to test the blimp robot's ability to navigate and maintain stability in turbulent environments. The total duration of the journey, including takeoff and landing, is 45 seconds. During the flight, the blimp robot reaches and maintains a height of approximately 4 meters, which introduces a slight delay during landing. The turbulence encountered during the flight affects the trajectory of the blimp, causing it to take detours. However, despite the challenging conditions, the blimp successfully overcomes the wind, maintains stability, and ultimately reaches the intended destination point.
\begin{figure}[t]
    \centering
    \includegraphics[width=0.9\linewidth]{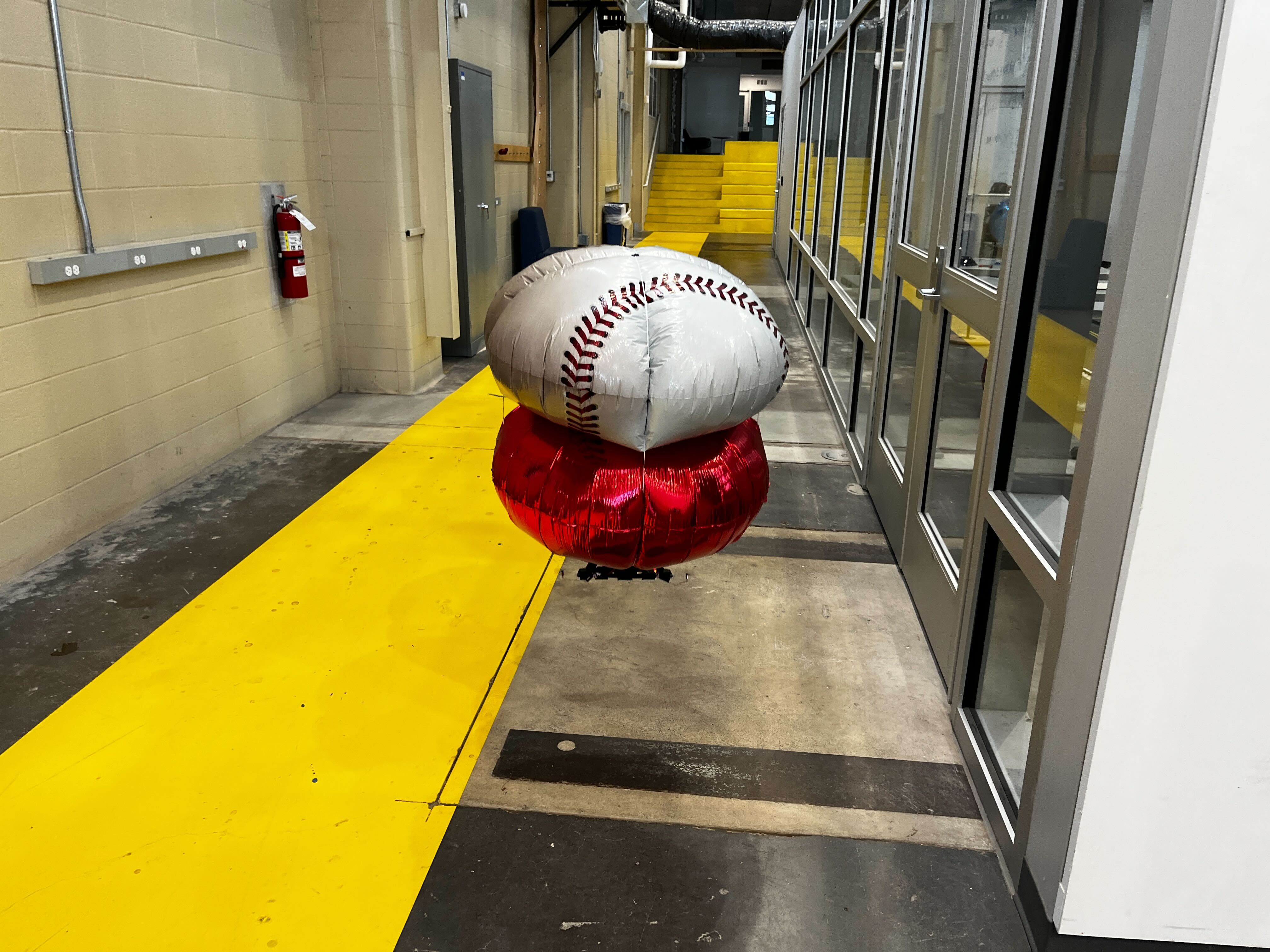}
    \caption{A blimp robot carrying an object weighing $50 \ g$ flying in a cluttered, narrow hallway with the presence of various obstacles including wall, trash cans, desks, and helmets hanging on the wall.}
    \label{fig:narrow}
    \vspace{-0.7em}
\end{figure}
\subsection{Flying in a cluttered, narrow space}
In this experiment, our aim is to showcase the collision-tolerant capabilities of our blimp robot. We design the experiment to test the blimp's ability to passively withstand collisions by intentionally directing it towards different obstacles along its path. We conduct the experiment in a narrow hallway with obstacles of various shapes and sizes (See Fig. 7). During the experiment, the blimp robot not only tolerates hard crashes, but it also effectively protects its propellers from hitting the obstacles, preventing damages to the vehicle. Despite the collisions, the blimp robot exhibits the ability to recover quickly, maintaining its stability, and continue to proceed. This experiment highlights the robustness and durability of our blimp robot design, demonstrating its potential to operate in cluttered environments where collisions with obstacles are likely to occur.
\smallskip

\section{Conclusions and Future Work}
While blimp robots utilizing affordable components offer several advantages over traditional drones, such as extended flight time, improved payload capacity, and improved collision tolerance, and are a more cost-effective option for accomplishing similar tasks, there are still several challenges that must be addressed to enable their seamless use and widespread adoption in projects aimed at generating significant social impact, such as improving accessibility in rural communities. In light of that, the authors of this paper are actively working as part of a team to develop a cost-effective and reliable prototype capable of autonomous flight in outdoor environments.
\smallskip

The current design of the blimp robot lacks autonomy or a control strategy beyond the remote controller. Incorporating additional sensors such as Geo-localization sensors or cameras into the system becomes crucial to address the potential impact of the added weight on the blimp's overall performance. Adding extra weight will necessitate generating more uplift force to uphold stable flight. Therefore, it will be essential to make proper adjustments and optimize the blimp's structure and propulsion system to accommodate the increased payload while ensuring its effective maneuver and stability.
\smallskip



Moving forward, we are strongly committed to integrating a perception component into our blimp robot. We are considering the utilization of potential camera modules, such as OpenMV H7 and Nicla Vision, both of which offer open-source capabilities and high versatility. The integration of perception capabilities involves the implementation of visual servoing methodologies, such as path following, which has been widely adopted as a prominent control strategy for robot blimps. With the integration of the path following algorithm, the blimp will be empowered to navigate autonomously and accurately follow a specific trajectory, enhancing its capacity to efficiently and precisely connect two points in a straight line. The inclusion of perception features not only elevates the blimp's autonomy but also expands its potential for sophisticated tasks, showcasing its adaptability and utility in real-world applications. 
\smallskip

The current estimated vehicle cost is \$27.42 per vehicle, with a target to reduce it below \$150.00 after integrating the perception component.  While cost minimization is a priority, we maintain a focus on vehicle reliability and usability. One significant challenge we are addressing is the impact of wind on the blimp's safe operation. Wind can affect stability and control, potentially leading to undesired flight behavior. The 9-axis inertial motion sensors, which accurately monitor the vehicle's orientation, are currently incorporated to enable the vehicle's effective response during turbulence and to maintain stable flight. Moreover, we are actively exploring and designing structural elements that enhance wind tolerance. These modifications aim to make the vehicle more resilient to gusts and crosswinds.
\smallskip

In our future endeavors, we aim to improve accessibility and provide timely assistance to communities in need. Our project envisions the establishment of a sustainable and scalable blimp robot that serves as a reliable means of transportation for critical supplies and medical assistance. To ensure the successful implementation, the team has actively sought continuous feedback from partners in the Philippines. During our initial visit to the Philippines in July 2023, we formed essential partner relationships to establish developmental frameworks for the vehicle. Key partnerships were forged with the Red Cross Philippines and the Quezon City Disaster and Risk Reduction Management Office. These collaborations have significantly contributed to narrowing the focus of our project, aligning it with the needs and priorities of the local communities.

\smallskip
Moving forward, our goal is to develop a vehicle that effectively addresses a well-defined problem. We plan to thoroughly test and implement this vehicle on a small scale, ensuring its functionality and adaptability to local conditions. Upon a successful small-scale launch, we aim to pursue widespread implementation of the blimp robot throughout the Philippines. Aligned with our vision, we are committed to generating educational impact and promoting knowledge transfer. We aim to make technology more affordable and accessible. By providing training and educational programs, we can empower local communities to effectively and sustainably utilize and maintain the blimp robot rescue network.

\smallskip



\bibliographystyle{IEEEtran}
\bibliography{references}

\end{document}